\documentclass[10pt, conference]{IEEEtran}
\IEEEoverridecommandlockouts
\usepackage{xcolor}        
\usepackage{graphicx}
\usepackage{amsmath, amssymb, amsfonts, amsthm}
\usepackage[caption=false, font=footnotesize]{subfig}
\usepackage{textcomp}
\usepackage{cite}
\usepackage{multirow}
\usepackage[ruled,vlined,linesnumbered]{algorithm2e}
\DeclareMathOperator*{\argmax}{arg\,max}

\DeclareMathOperator*{\maximize}{maximize}

\DeclareMathOperator{\tr}{tr}

\def\BibTeX{{\rm B\kern-.05em{\sc i\kern-.025em b}\kern-.08em
    T\kern-.1667em\lower.7ex\hbox{E}\kern-.125emX}}
\begin{document}

\title{Agentic AI for Intent-driven Optimization in Cell-free O-RAN\\}
\author{\IEEEauthorblockN{Mohammad Hossein Shokouhi and Vincent W.S. Wong}
\IEEEauthorblockA{Department of Electrical and Computer Engineering, The University of British Columbia, Vancouver, Canada \\
email: \{mhshokouhi, vincentw\}@ece.ubc.ca}
}


\maketitle
\begin{abstract}
Agentic artificial intelligence (AI) is emerging as a key enabler for autonomous radio access networks (RANs), where multiple large language model (LLM)-based agents reason and collaborate to achieve operator-defined intents.
The open RAN (O-RAN) architecture enables the deployment and coordination of such agents.
However, most existing works consider simple intents handled by independent agents, while complex intents that require coordination among agents remain unexplored.
In this paper, we propose an agentic AI framework for intent translation and optimization in cell-free O-RAN. 
A supervisor agent translates the operator intents into an optimization objective and minimum rate requirements. 
Based on this information, a user weighting agent retrieves relevant prior experience from a memory module to determine the user priority weights for precoding. If the intent includes an energy-saving objective, then an open radio unit (O-RU) management agent will also be activated to determine the set of active O-RUs by using a deep reinforcement learning (DRL) algorithm. 
A monitoring agent measures and monitors the user data rates and coordinates with other agents to guarantee the minimum rate requirements are satisfied.
To enhance scalability, we adopt a parameter-efficient fine-tuning (PEFT) method that enables the same underlying LLM to be used for different agents.
Simulation results show that the proposed agentic AI framework reduces the number of active O-RUs by $41.93\%$ when compared with three baseline schemes in energy-saving mode. Using the PEFT method, the proposed framework reduces the memory usage by $92\%$ when compared with deploying separate LLM agents.
\end{abstract}

\section{Introduction}
The open radio access network (O-RAN) architecture has emerged as a key enabler of the sixth-generation (6G) mobile wireless networks \cite{survey:O-RAN1}. 
O-RAN disaggregates the traditional base station into open radio units (O-RUs), open distributed units (O-DUs), and open central units (O-CUs). O-DUs host distributed applications (dApps) that execute real-time (sub-$10$ ms) control logic \cite{melodia:dApp_full}. 
O-RAN also introduces two RAN intelligent controllers (RICs).
The near-real-time (near-RT) RIC hosts xApps that perform near-RT ($10$ ms to $1$ s) control tasks. 
The non-RT RIC hosts rApps that perform non-RT (beyond $1$ s) tasks \cite{survey:O-RAN1}. 
Together, rApps, xApps, and dApps enable intelligent control of RAN across different timescales.
For instance, in \cite{melodia:xApp-scheduler}, an xApp adjusts the weights of users in a proportional fair scheduler to guarantee the minimum data rate for each user. 
In \cite{INFOCOM_ORAN1}, an O-RU on/off controller rApp and an O-CU/O-DU function placement xApp are proposed to minimize the energy consumption in O-RAN.

By leveraging the capabilities of the O-RAN architecture, researchers are now advancing the vision of autonomous RANs that use artificial intelligence (AI) to adapt and optimize without human intervention. A key enabler of this vision is agentic AI, where multiple specialized AI agents collaborate to monitor, analyze, and control the RAN. Agentic AI provides human-like reasoning and adaptability while operating at machine speed and scale.
Traditional RAN management requires operators to navigate complex dashboards and maintain expertise across diverse systems. 
In contrast, AI-enabled RANs allow operators to express high-level intents or long-term goals in natural language. The AI agents translate these directives into sub-tasks, configure the RAN, and monitor key performance indicators to achieve the desired outcome. 
Large language models (LLMs), with their exceptional natural language understanding, reasoning, and generalization capabilities, are promising enablers of agentic AI.

Recent works have proposed various AI agents for O-RAN optimization. 
In \cite{agentic_AI:ALLSTaR}, LLM agents interpret the operator intents and match the intents to the most suitable scheduler. 
In \cite{agentic_AI:MX-AI}, an agent decomposes the operator intents into sub-tasks and invokes specialized agents to accomplish the objective.
In \cite{agentic_AI:AgentRAN}, a manager agent in the non-RT RIC decomposes the operator intents into sub-intents for power control and resource allocation agents in the near-RT RIC. 
The aforementioned works \cite{agentic_AI:ALLSTaR, agentic_AI:MX-AI, agentic_AI:AgentRAN} only consider the scenario where the operator intents can be decomposed into non-overlapping objectives handled by independent agents. The case of complex intents that require inter-agent collaboration remains largely unexplored. Moreover, most works deploy a separate LLM for each agent, which may limit the scalability.

In this paper, we propose an agentic AI framework for intent translation and optimization in cell-free O-RAN, where each user is collaboratively served by multiple O-RUs. LLM-based agents translate the operator intents, which are expressed in natural language, into specific objectives and jointly determine the control parameters that achieve the desired outcome. The contributions of this paper are as follows:
\begin{itemize}
    \item We deploy multiple agents that operate at different timescales and interact through standardized interfaces. A supervisor agent in the non-RT RIC translates the operator intent into an objective function and minimum rate requirements. The objective function can be a utility function related to the data rates or energy saving. Based on this information, the user weighting agent in the near-RT RIC retrieves relevant prior experience from a memory module to determine the user priority weights for precoding. In the energy-saving mode, the O-RU management agent invokes a multi-agent deep reinforcement learning (DRL) algorithm to determine the set of active O-RUs. The O-DU uses these parameters for precoding.
    \item The user data rates depend on both the user priority weights and the set of active O-RUs. Inter-agent collaboration is required to guarantee the minimum data rate requirements. 
    To this end, a monitoring agent measures and monitors the user data rates, coordinates the user weighting and O-RU management agents, and provides feedback to them until the minimum data rate constraints are satisfied.
    \item Due to the substantial size of LLMs, deploying separate LLM agents in the near-RT RIC limits scalability. To address this issue, we deploy a shared LLM in the near-RT RIC and train a quantized low-rank adaptation (QLoRA) adapter for each agent. This approach preserves agent specialization while greatly reducing memory usage.
    \item Simulation results show that the proposed agentic AI framework in energy-saving mode reduces the number of active O-RUs by $41.93\%$ when compared with three baseline schemes. Furthermore, using QLoRA, the proposed framework reduces the memory usage by $92\%$ when compared with deploying separate LLM agents.
\end{itemize}

\begin{figure}[t]
\center{\includegraphics[width=0.8\linewidth]{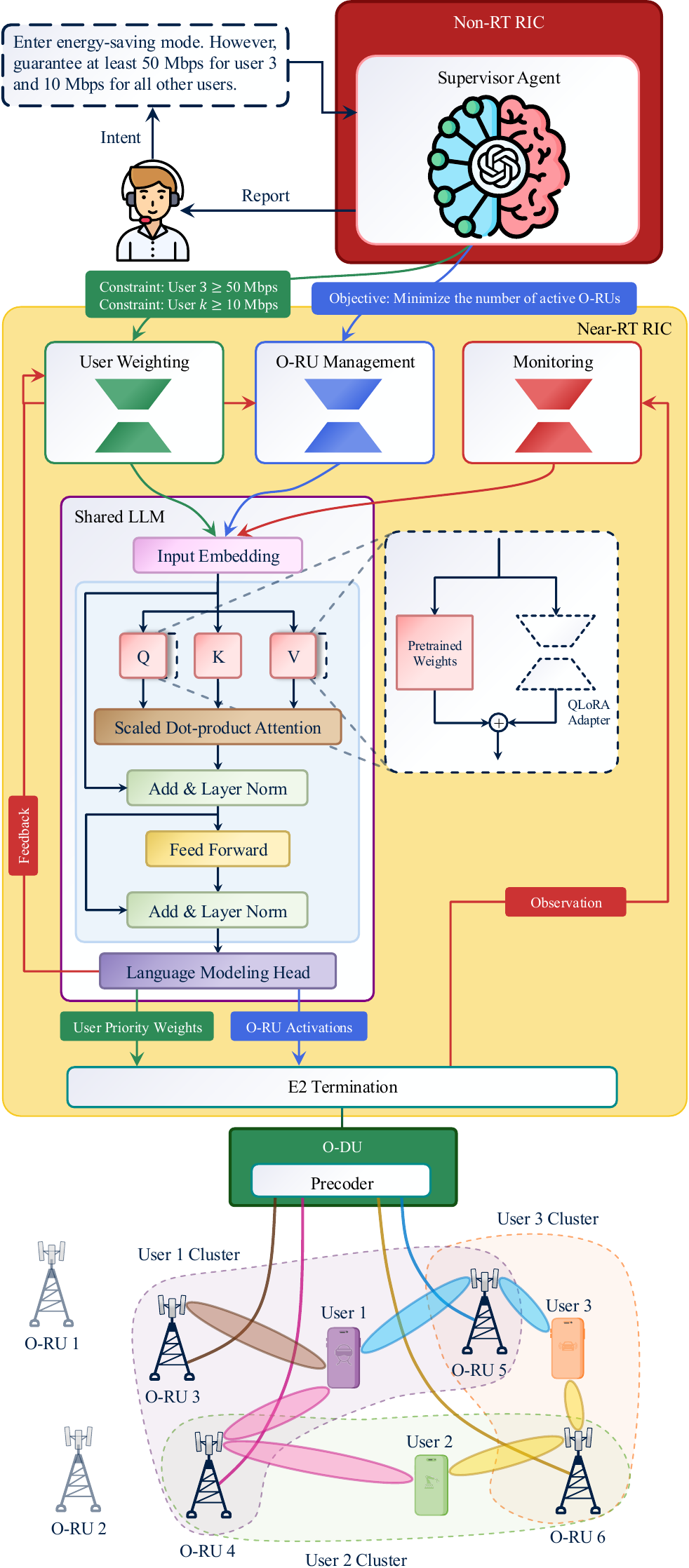}}
\caption{The considered system model. The operator intents are translated into objectives by the supervisor agent in the non-RT RIC. Near-RT agents determine the user priority weights and the set of active O-RUs. Agents in the near-RT RIC share an LLM with different QLoRA adapters.}
\label{fig:model}
\vspace{-5mm}
\end{figure}

The rest of this paper is organized as follows. In Section \ref{sec:model}, we introduce the system model and formulate the optimization problem. In Section \ref{sec:algorithm}, we present the proposed framework and describe the role of each agent and the workflow from intent translation to optimization. Performance evaluation is presented in Section \ref{sec:eval}. Conclusion is given in Section \ref{sec:conclusion}.

{\it Notations}: In this paper, $\mathbb{C}$ and $\mathbb{R}$ denote the set of complex and real numbers, respectively. Boldface uppercase letters (e.g., $\mathbf{X}$) represent matrices, while boldface lowercase letters (e.g., $\mathbf{x}$) represent vectors. The $N \times N$ identity matrix is denoted by $\mathbf{I}_N$. $(\cdot)^\mathrm{H}$ denotes the conjugate transpose of a vector or matrix. For a matrix, $\tr(\cdot)$ and $\det(\cdot)$ denote the trace and determinant, respectively. $\langle \mathbf{x}, \mathbf{y} \rangle$ denotes the inner product of vectors $\mathbf{x}$ and $\mathbf{y}$.

\section{System Model and Problem Formulation}\label{sec:model}
We consider downlink transmission in a cell-free O-RAN as shown in Fig. \ref{fig:model}. The set of users is denoted by $\mathcal{K}=\{1, 2, \ldots, K\}$. 
The set of O-RUs is denoted by $\mathcal{L}=\{1,2,\ldots,L\}$. Each O-RU has $N_\textrm{t}$ transmit antennas. Each user equipment has $N_\textrm{r}$ receive antennas. 
The O-DU is connected to O-RUs via open fronthaul (O-FH) links. 

In cell-free O-RAN, each user $k \in \mathcal{K}$ is coherently served by a subset of O-RUs, denoted by $\mathcal{L}_k \subset \mathcal{L}$. This results in a more uniform throughput across the coverage area compared to the cellular architecture. The subset of users served by O-RU $l \in \mathcal{L}$ is denoted by $\mathcal{K}_l \subset \mathcal{K}$. The operator may deactivate certain O-RUs during low-traffic periods for energy saving. Let $z_l \in \{0,1\}$ denote the O-RU activation variable, where $z_l=1$ if O-RU $l$ is active and $z_l=0$ otherwise. Let $\beta_{k,l}$ be the large-scale fading coefficient between user $k$ and O-RU $l$. 
We use a heuristic \cite{MIMO:MHSH2} where each user $k$ can be served by at most $L^\textrm{max}$ active O-RUs which have the largest $\beta_{k,l}$.

Let $\mathbf{H}_{k,l} \in \mathbb{C}^{N_\textrm{r} \times N_\textrm{t}}$ denote the downlink channel matrix between user $k \in \mathcal{K}$ and O-RU $l \in \mathcal{L}$.
Let $\mathbf{V}_{k,l} \in \mathbb{C}^{N_\textrm{t} \times N_\textrm{s}}$ denote the precoding matrix at O-RU $l$ for user $k \in \mathcal{K}_l$, where $N_\textrm{s}$ is the number of data streams.
Let $\mathbf{\Gamma}_{k} \in \mathbb{C}^{N_\textrm{r} \times N_\textrm{r}}$ denote the signal-to-interference-plus-noise ratio (SINR) matrix of user $k$. It can be expressed as \cite{MIMO:MHSH1}
\begin{equation}\label{eq:SINR}
    \mathbf{\Gamma}_{k} = \mathbf{\Psi}_{k,k}\mathbf{\Psi}_{k,k}^\textrm{H} \left(\sum_{i \in \mathcal{K} \setminus \{k\}} \mathbf{\Psi}_{k,i}\mathbf{\Psi}_{k,i}^\textrm{H} + \sigma^2 \mathbf{I}_{N_\textrm{r}} \right)^{-1},
\end{equation}
where $\sigma^2$ is the variance of the additive white Gaussian noise, and $\mathbf{\Psi}_{k,i} = \sum_{l \in \mathcal{L}_i} \mathbf{H}_{k,l} \mathbf{V}_{i,l} \in \mathbb{C}^{N_\textrm{r} \times N_\textrm{s}}$ denotes the effective channel matrix for user $k$ if $i=k$, and the effective interference matrix from user $i$ to user $k$ otherwise.
The achievable data rate of user $k \in \mathcal{K}$ is given by $r_{k} = \log_2 \det \left(\mathbf{I}_{N_\textrm{r}} + \mathbf{\Gamma}_{k}\right)$.

The optimization problem aims to maximize a utility function while satisfying the minimum rate requirements of users and the maximum transmit power at each O-RU. We have
\begin{subequations} \label{eq:optimization}
\begin{alignat}{2}
&\maximize\limits_{\substack{\mathbf{V}_{k,l}, z_l\\ k \in \mathcal{K}, l \in \mathcal{L}}} & &\quad U(\mathbf{V},\mathbf{z}) \label{eq:obj}\\
&\textrm{subject to} & &\quad r_{k} \geq R_{k}^{\textrm{min}}, \; k \in \mathcal{K} \label{eq:cons1}\\
& & &\quad \sum_{k \in \mathcal{K}_l} \tr\left(\mathbf{V}_{k,l}\mathbf{V}_{k,l}^\textrm{H}\right) \leq z_l P^{\textrm{max}}, \; l \in \mathcal{L} \label{eq:cons2}\\
& & &\quad  z_l \in \{0,1\}, \; l \in \mathcal{L},\label{eq:cons3}
\end{alignat}
\end{subequations}
where $U(\cdot)$ is the utility function. $\mathbf{V} = \left[\mathbf{V}_{k,l},\;\forall k \in \mathcal{K}, l \in \mathcal{L}\right]$ is the stacked precoding matrix. $\mathbf{z} = \left[z_l,\;\forall l \in \mathcal{L}\right]$ is the stacked O-RU activation vector.
Constraint \eqref{eq:cons1} guarantees the minimum data rate requirement $R_{k}^{\textrm{min}}$ of each user $k$. Constraint \eqref{eq:cons2} enforces the transmit power limit at each O-RU. If O-RU $l$ is active, i.e., $z_l=1$, then its total transmit power must not exceed $P^{\textrm{max}}$. If O-RU $l$ is inactive, i.e., $z_l=0$, then its transmit power is set to zero.

We consider two types of objective functions. The first type corresponds to the aggregate utility, defined as $U=\sum_{k \in \mathcal{K}} U_k(r_k)$, where $U_k(r_k)$ is a concave utility function. The second type targets energy savings, defined as $U=-\sum_{l \in \mathcal{L}} z_l$, which aims to minimize the number of active O-RUs. In both cases, problem \eqref{eq:optimization} is nonconvex due to the nonconvexity of the objective function \eqref{eq:obj} and constraint \eqref{eq:cons1}.

\section{Agentic AI Framework for Cell-free O-RAN}\label{sec:algorithm}
In this section, we first propose algorithms to solve problem \eqref{eq:optimization} for both types of objective functions. 
In the case of aggregate-utility maximization, the number of active O-RUs does not appear in the objective function. Therefore, we keep all the O-RUs active, i.e., $z_l=1,\; l \in \mathcal{L}$, and let the precoding algorithm determine the power allocation across O-RUs to maximize the aggregate utility. Using the method of Lagrange multipliers, constraint \eqref{eq:cons1} can be incorporated into the objective function \eqref{eq:obj}. 
Constraint \eqref{eq:cons2} remains as an explicit constraint in the dual problem. The partial Lagrange dual function is $g(\boldsymbol{\mu}) = \sup_{\mathbf{V} \in \mathcal{D}} \sum_{k \in \mathcal{K}} \left(U_k(r_k) + \mu_k \left(r_{k} - R_{k}^{\textrm{min}}\right)\right)$,
where $\boldsymbol{\mu}= \left[\mu_k,\;\forall k \in \mathcal{K}\right]^\top$ is the vector of Lagrange multipliers, and $\mathcal{D}=\left\{\mathbf{V}: \sum_{k \in \mathcal{K}_l} \tr\left(\mathbf{V}_{k,l}\mathbf{V}_{k,l}^\textrm{H}\right) \leq P^{\textrm{max}}, l \in \mathcal{L}\right\}$. 
To obtain the dual function, we need to solve the inner supremum over $\mathbf{V}$. Let $\tilde{U}_k(r_k)=U_k(r_k) + \mu_k \left(r_{k} - R_{k}^{\textrm{min}}\right)$. Thus, we have
\begin{align}
&\maximize\limits_{\substack{\mathbf{V}_{k,l},\\ k \in \mathcal{K}_l, l \in \mathcal{L}}} \quad\sum_{k \in \mathcal{K}} \tilde{U}_k \label{eq:inner_supremum}\\
&\textrm{subject to constraint \eqref{eq:cons2}}. \notag
\end{align}
The Lagrange dual problem can be expressed as
\begin{subequations} \label{eq:dual_problem}
\begin{alignat}{2}
&\inf_{\mu_k, k \in \mathcal{K}} & &\quad g(\boldsymbol{\mu}) \label{eq:dual_obj}\\
&\textrm{subject to} & &\quad \mu_k \geq 0, \; k \in \mathcal{K}. \label{eq:dual_cons}
\end{alignat}
\end{subequations}
We use the dual gradient ascent method to iteratively solve subproblems \eqref{eq:inner_supremum} and \eqref{eq:dual_problem} for $\mathbf{V}$ and $\boldsymbol{\mu}$, respectively.
Problem \eqref{eq:inner_supremum} can be reformulated as an equivalent weighted minimum mean square error (WMMSE) problem by introducing the auxiliary weight matrix $\mathbf{W}_k \in \mathbb{C}^{N_\textrm{s}\times N_\textrm{s}}$ and receive filter matrix $\mathbf{U}_k \in \mathbb{C}^{N_\textrm{r}\times N_\textrm{s}}$ \cite{MIMO1}. The precoding matrices can be determined by iteratively updating $\mathbf{W}_k$, $\mathbf{U}_k$, and $\mathbf{V}_{k,l}$ as \cite{MIMO:MHSH2}
\begin{equation}\label{eq:U_WMMSE}
    \mathbf{U}_k = \left(\sum_{i \in \mathcal{K}} \mathbf{\Psi}_{k,i} \mathbf{\Psi}_{k,i}^\textrm{H} + \sigma^2 \mathbf{I}_{N_\textrm{r}}\right)^{-1} \mathbf{\Psi}_{k,k},
\end{equation}
\vspace{0.1mm}
\begin{equation}\label{eq:W_WMMSE}
    \mathbf{W}_k = \left(\mathbf{I}_{N_\textrm{s}} - \mathbf{U}_k^\textrm{H} \mathbf{\Psi}_{k,k} \right)^{-1},
\end{equation}
\begin{align}\label{eq:V_WMMSE}
    \mathbf{V}_{k,l} = &\left(\sum_{i \in \mathcal{K}_l}\alpha_i \mathbf{H}_{i,l}^\textrm{H} \mathbf{X}_i \mathbf{H}_{i,l} + \xi_l \mathbf{I}_{N_\textrm{t}}\right)^{-1} \nonumber\\ &\left(\alpha_k\mathbf{H}_{k,l}^\textrm{H} \mathbf{Y}_k^\textrm{H} - \sum_{i \in \mathcal{K}_l} \alpha_i \mathbf{H}_{i,l}^\textrm{H} \mathbf{X}_i^\textrm{H} \mathbf{Z}_{i,k,l}\right),
\end{align}
where $\mathbf{X}_k = \mathbf{U}_k \mathbf{W}_k \mathbf{U}_k^\textrm{H}$, $\mathbf{Y}_k = \mathbf{W}_k \mathbf{U}_k^\textrm{H}$, $\mathbf{Z}_{i,k,l} = \sum_{j \in \mathcal{L}_k \setminus\{l\}} \mathbf{H}_{i,j} \mathbf{V}_{k,j}$, and $\xi_l$ is the Lagrange multiplier that must satisfy the complementary slackness condition of constraint \eqref{eq:cons2}. $\alpha_k$ is the priority weight of user $k$, given by \cite{MIMO1}
\begin{equation}\label{eq:priority}
    \alpha_k = \tilde{U}'_k(r_k) = U'_k(r_k) + \mu_k,
\end{equation}
where $\tilde{U}'_k(r_k)$ is the derivative of $\tilde{U}_k(r_k)$ with respect to $r_k$. Finally, $\mu_k$ can be updated for each user $k$ in each iteration by using gradient ascent as $\mu_k \xleftarrow{} \mu_k + \zeta_k \left( R_{k}^{\textrm{min}} - r_{k} \right)$, where $\zeta_k$ is the step size for user $k$. 

In the case of energy savings with $U=-\sum_{l \in \mathcal{L}} z_l$, problem \eqref{eq:optimization} becomes a mixed-integer programming problem, which is NP-hard. We use the block coordinate descent method to solve the problem for $\mathbf{V}_{k,l}$ and $z_l$ iteratively. 
We can still use \eqref{eq:U_WMMSE}, \eqref{eq:W_WMMSE}, and \eqref{eq:V_WMMSE} to find a solution to the precoding subproblem.
However, the O-RU activation subproblem is combinatorial and hard to solve. Due the large number of O-RUs in cell-free O-RAN, centralized optimization algorithms may lead to scalability challenges and slow convergence.

To solve the O-RU activation subproblem in a distributed manner, we use the multi-agent proximal policy optimization (MAPPO) DRL algorithm \cite{MAPPO}.
MAPPO allows each agent to learn its own policy while coordinating through a shared critic.

Each O-RU $l$ is modeled as an agent that observes its local environment and decides its activation state.
Each near-RT loop is treated as one step of the DRL algorithm.
The action of agent $l$ at step $t$ represents whether O-RU $l$ is active or inactive, i.e., $a_l^{(t)} \triangleq z_l^{(t)} \in \{0,1\}$.
All agents share the same reward function, which is defined as
\begin{equation}\label{eq:overall_reward}
    R^{(t)} = - \frac{1}{L}\sum_{l \in \mathcal{L}} z_l^{(t)} - \frac{1}{K} \sum_{k \in \mathcal{K}} \lambda_k \upsilon_k^{(t)} - \frac{1}{L} \sum_{l \in \mathcal{L}} \left|z_l^{(t)} - z_l^{(t-1)}\right|,
\end{equation}
where the first term penalizes the number of active O-RUs. $\upsilon_k^{(t)}=\max\left\{0,R_k^\textrm{min}-r_k^{(t)}\right\}$ denotes the data rate violation of user $k$, and $\lambda_k$ is the violation penalty coefficient. The third term discourages frequent changes in O-RU activation.
Consequently, the local observation of agent $l$ includes the large-scale fading coefficients between O-RU $l$ and all users, the effective rate violations of users, and the previous activation state of O-RU $l$. It is defined as $o_l^{(t)}=\left[\left[\log(\beta_{k,l}),\tanh \left(\lambda_k \upsilon_k^{(t)}\right)\right]_{k \in \mathcal{K}}, z_l^{(t-1)}\right]$.

At each step $t$, each agent $l$ makes the local observation $o_l^{(t)}$ and selects an action $a_l^{(t)} \sim \pi_{\theta_l}\left(\cdot \mid o_l^{(t)}\right)$ according to its policy $\pi_{\theta_l}$ with parameters $\theta_l$. The agents then receive a reward $R^{(t)}$ and the environment transitions to the next state.

To evaluate the quality of the joint action $\mathbf{a}^{(t)}=\left(a_1^{(t)},\ldots,a_L^{(t)}\right)$ at step $t$, MAPPO uses the advantage function. It is expressed as
\begin{equation}\label{eq:advantage}
    \hat{A}^{(t)}=\sum_{i=0}^{T-t-1} \gamma^i \left(R^{(t+i)}+\gamma V_{\phi}\left(\mathbf{o}^{(t+i+1)}\right) - V_{\phi}\left(\mathbf{o}^{(t+i)}\right)\right),
\end{equation}
where $\gamma$ is the discount factor and $T$ is the episode length. $\mathbf{o}^{(t)}=\left(o_1^{(t)},\ldots,o_L^{(t)}\right)$ is the joint observations of agents and $V_\phi(\mathbf{o}^{(t)})$ is the state-value function estimated by a critic with parameters $\phi$.
The loss function of agent $l$ is defined as
\begin{equation}
    L_l(\theta_l, \phi) = -L^\textrm{CLIP}(\theta_l) - c_1 L^\textrm{ENT}(\theta_l) + c_2 L^\textrm{V}(\phi),
\end{equation}
where $c_1$ and $c_2$ are constant weights, and
\begin{multline}\label{eq:L_CLIP}
    L^\textrm{CLIP}(\theta_l) = \hat{\mathbb{E}}_t \Big[
    \min\Big(
    \rho^{(t)} (\theta_l) \hat{A}^{(t)},\\
    \textrm{clip} \left(
        \rho^{(t)} (\theta_l) , 1-\epsilon, 1+\epsilon
    \right)\hat{A}^{(t)}
    \Big)
    \Big],
\end{multline}
\begin{equation}\label{eq:L_ENT}
    L^\textrm{ENT}(\theta_l) = \hat{\mathbb{E}}_t \left[
        \mathcal{H}\left(\pi_{\theta_l}\left(\cdot \mid o_l^{(t)}\right)\right)
    \right],
\end{equation}
\begin{equation}\label{eq:L_V}
    L^\textrm{V}(\phi) = \hat{\mathbb{E}}_t \left[
        \left(V_\phi\left(\mathbf{o}^{(t)}\right)-\hat{G}^{(t)}\right)^2
    \right].
\end{equation}
In \eqref{eq:L_CLIP}, $\hat{\mathbb{E}}_t[\cdot]$ is the empirical average over the collected samples in the past episode. $\rho^{(t)} (\theta_l) = \frac{\pi_{\theta_l}\left(a_l^{(t)}\mid o_l^{(t)}\right)}{\pi_{\theta_l}^\textrm{old}\left(a_l^{(t)} \mid o_l^{(t)}\right)}$ denotes the probability ratio between the new and old policies, and $\epsilon$ controls the clipping range. The function $L^\textrm{CLIP}$ encourages actions with higher advantages while preventing large policy updates by restricting $\rho^{(t)} (\theta_l)$ within $[1-\epsilon,1+\epsilon]$. In \eqref{eq:L_ENT}, $\mathcal{H}\left(\pi_{\theta_l}\left(\cdot \mid  o_l^{(t)}\right)\right)=-\sum_{a_l} \pi_{\theta_l}\left(a_l \mid o_l^{(t)}\right) \log \pi_{\theta_l}\left(a_l \mid o_l^{(t)}\right)$ represents the policy entropy. $L^\textrm{ENT}$ encourages exploration by keeping the policy stochastic. The function \eqref{eq:L_V} represents the state-value estimation error, where $\hat{G}^{(t)}=\sum_{i=0}^{T-t-1}\gamma^i R^{(t+i)}$ is the target return at step $t$.

During training, the trajectory $\tau=\left\{\left(\mathbf{o}^{(t)},\mathbf{a}^{(t)},R^{(t)},\mathbf{o}^{(t+1)}\right)\right\}_{t=0}^{T-1}$ is collected in each episode under the current joint policy $\boldsymbol{\pi}=\left(\pi_{\theta_1},\ldots,\pi_{\theta_L}\right)$. The advantage function $\hat{A}^{(t)}$ is then estimated for each step using \eqref{eq:advantage}. Then, the policy parameters $\theta_l$ of each agent $l$ are updated via $\theta_l \leftarrow \theta_l + \delta_l \nabla_{\theta_l}L_l$, where $\delta_l$ denotes the learning rate of the policy. The critic parameters are updated using $\phi \leftarrow \phi - \delta_v \nabla_{\phi}L^\textrm{V}$, where $\delta_v$ is the learning rate of the critic. After training, each O-RU $l$ independently determines its activation state $z_l^{(t)}$ based on its local observation $o_l^{(t)}$ and the learned policy $\pi_{\theta_l}$. Algorithm 1 summarizes one iteration of training and inference in the proposed DRL algorithm.

\addtolength{\topmargin}{+0.01in}
\begin{algorithm}[t]\small
    \caption{The proposed multi-agent DRL algorithm for O-RU activation management} \label{alg:DRL}
    \textit{Training:}\\
    \For{$t:=0$ \textnormal{to} $T - 1$}{
        For each agent $l$, sample an action $a_l^{(t)} \sim \pi_{\theta_l}\left(\cdot \mid o_l^{(t)}\right)$\\
        Execute the joint action $\mathbf{a}^{(t)}$ and observe the shared reward $R^{(t)}$ and next joint observation $\mathbf{o}^{(t+1)}$\\
        Store $\left(\mathbf{o}^{(t)},\mathbf{a}^{(t)},R^{(t)},\mathbf{o}^{(t+1)}\right)$ in the trajectory $\tau$\\
    }
    For each step $t$ of the trajectory $\tau$, estimate $\hat{A}^{(t)}$ via \eqref{eq:advantage}\\
    For each agent $l$, update the policy as $\theta_l \leftarrow \theta_l + \delta_l \nabla_{\theta_l}L_l$\\
    Update the centralized critic as $\phi \leftarrow \phi - \delta_v \nabla_{\phi}L^\textrm{V}$\\
    \textit{Inference:}\\
    For each agent $l$, select $a_l^{(t)} := \argmax_{a} \pi_{\theta_l}\left(a \mid o_l^{(t)}\right)$
\end{algorithm}
\setlength{\textfloatsep}{0pt}

Note that the violation penalty coefficient $\lambda_k$ in \eqref{eq:overall_reward} affects which O-RUs are activated by the DRL algorithm. Thus, both the priority weight $\alpha_k$ and the violation penalty coefficient $\lambda_k$ influence user data rate $r_k$, and their updates must be coordinated to satisfy the minimum rate requirements and prevent oscillations. For instance, if these coefficients are updated independently, $\lambda_k$ may increase rapidly before the user priority weights $\alpha_k$ converge, which forces the DRL algorithm to activate unnecessary O-RUs. To enable intent translation and coordination, we deploy the following LLM agents: A supervisor agent is deployed as an rApp in the non-RT RIC. The user weighting, O-RU management, and monitoring agents are deployed as xApps in the near-RT RIC. 

The supervisor agent receives the operator intent, translates it into specific objectives, and forwards them to the near-RT agents over the A1 interface. An example of the operator intent and the extracted objectives is shown in Fig. \ref{fig:model}.

The user weighting agent receives a prompt that includes the utility functions $U_k(\cdot)$, the minimum rate requirements $R_k^\textrm{min}$, and the history of $[\upsilon_k, \mu_k]$ over a window of $W$ near-RT loops. It then computes the derivatives $U'_k(\cdot)$, updates the Lagrange multipliers $\mu_k$,  and determines the priority weights $\alpha_k$ according to \eqref{eq:priority}.

The O-RU management agent receives the energy-saving mode and the recent history of $[\upsilon_k, \lambda_k]$ over a window of $W$ near-RT loops. It then determines the O-RU activations. If energy saving is not requested, all O-RUs remain active, i.e., $z_l=1,\; l \in \mathcal{L}$. When energy saving is enabled, the agent first updates the violation penalty coefficients $\lambda_k$ and then invokes the DRL xApp to determine the set of active O-RUs. 
The user priority weights $\alpha_k$ and O-RU activations $z_l$ are forwarded to the precoding dApp over the E2 interface, which then determines the precoding matrices via \eqref{eq:V_WMMSE}.

\begin{figure}[t]
\center{\includegraphics[width=\linewidth]{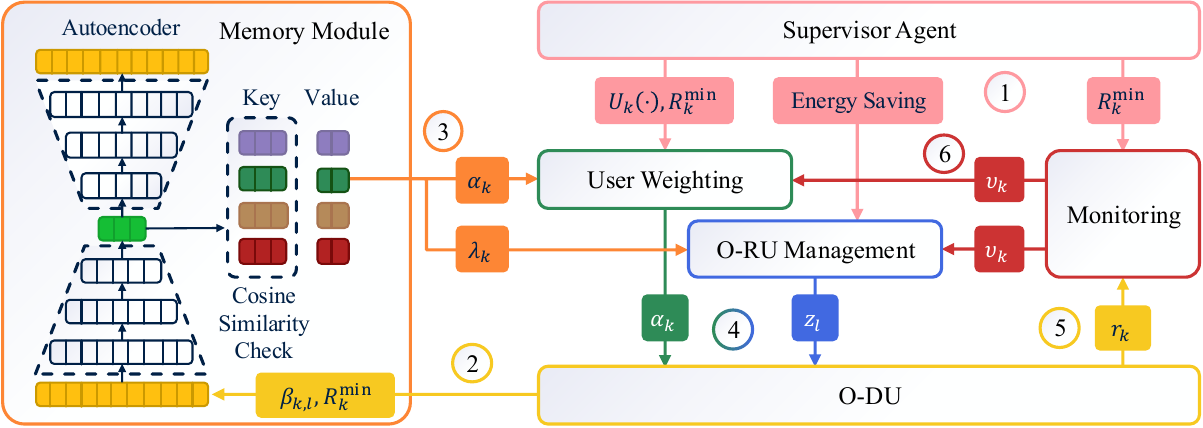}}
\caption{Block diagram of the proposed framework. The user weighting and O-RU management agents receive operator objectives from the supervisor agent, feedback from the monitoring agent, and prior knowledge from the memory module to determine the user weights and O-RU activations. Steps 4-6 are repeated until all the minimum rate requirements are satisfied.}
\label{fig:block_diagram}
\end{figure}

The monitoring agent continuously measures and monitors the user data rates and coordinates the user weighting and O-RU management agents to ensure that the minimum rate requirements are satisfied. It analyzes the recent history of $[\upsilon_k, \lambda_k, \alpha_k]$ over a window of $W$ near-RT loops and identifies users with violated rate constraints. For each violated user $k$, it decides whether to increase the user priority weight, $\alpha_k$. If $\alpha_k$ is already high, it instead informs the O-RU management agent to increase the violation penalty coefficient $\lambda_k$, which results in the activation of additional nearby O-RUs. This adjustment process continues until all constraints are satisfied.

To accelerate convergence, we propose a retrieval-augmented coefficient tuning method. Each time the coefficients $\alpha_k$ and $\lambda_k$ converge, a new key-value pair $(\mathbf{q}_i,\mathbf{y}_i)$ is stored in the memory set $\mathcal{M}$, where $i$ is the index of the stored experience. The value vector stores the learned coefficients as $\mathbf{y}_i=\left[\alpha_k, \lambda_k\right]_{k \in \mathcal{K}}$. The key represents the environment setup and is defined as $\mathbf{q}_i=\textrm{emb}\left(\left[[\beta_{k,l}]_{l \in \mathcal{L}}, R_k^\textrm{min}\right]_{k \in \mathcal{K}}\right)$, where $\textrm{emb}(\cdot)$ is a trained autoencoder that maps the input into a low-dimensional embedding. When a new operator intent arrives, the current environment features, $\beta_{k,l}$ and $R_k^\textrm{min}$, are embedded as $\mathbf{q}$. This embedding is then compared with all stored keys $\mathbf{q}_i \in \mathcal{M}$ using the cosine similarity metric $\frac{\langle\mathbf{q}_i , \mathbf{q}\rangle}{\|\mathbf{q}_i\| \|\mathbf{q}\|}$. The coefficients from the most similar entry are retrieved from the memory and provided to the user weighting and O-RU management agents as relevant prior experience. Fig. \ref{fig:block_diagram} shows the interactions among agents in the proposed framework.

To avoid the memory overhead of deploying multiple LLM agents, we use QLoRA \cite{QLoRA} to fine-tune a shared lightweight LLM in the near-RT RIC on different downstream tasks. In QLoRA, the pretrained model weights $\boldsymbol{\theta} \in \mathbb{R}^{d_\textrm{out}\times d_\textrm{in}}$ are first quantized using a 4-bit NormalFloat (FP4) quantization scheme to minimize memory usage. Let $\hat{\boldsymbol{\theta}}  \in \mathbb{R}^{d_\textrm{out}\times d_\textrm{in}}$ denote the quantized weights. Each near-RT agent introduces a rank-$d_\textrm{r}$ adapter in the form of two trainable matrices $\mathbf{A} \in \mathbb{R}^{d_\textrm{out}\times d_\textrm{r}}$ and $\mathbf{B} \in \mathbb{R}^{d_\textrm{r} \times d_\textrm{in}}$, where $d_\textrm{r} \ll \min \left(d_\textrm{in}, d_\textrm{out}\right)$. The effective weight then becomes $\boldsymbol{\theta}' = \hat{\boldsymbol{\theta}} + \frac{\eta}{d_\textrm{r}} \mathbf{A}\mathbf{B}$,
where $\eta$ controls the contribution of the low-rank adapter. Thus, the forward pass for a given input vector $\mathbf{x}$ can be expressed as
\begin{equation}\label{eq:forward_pass}
    \mathbf{y} = \boldsymbol{\theta}' \mathbf{x} = \hat{\boldsymbol{\theta}} \mathbf{x} + \frac{\eta}{d_\textrm{r}} \mathbf{A}(\mathbf{B} \mathbf{x}),
\end{equation}
where the first term represents the quantized backbone and the second term introduces agent-specific adjustments through the lightweight adapters. During fine-tuning, the quantized weights $\hat{\boldsymbol{\theta}}$ remain frozen and only the low-rank matrices $\mathbf{A}$ and $\mathbf{B}$ are updated. During inference, each near-RT agent loads its corresponding adapter and determines the output using \eqref{eq:forward_pass}.

To enable fine-tuning, a teacher model is prompted to serve as each near-RT agent, and its interactions are collected into datasets. Using these datasets, QLoRA adapters are trained on a lightweight student model for each near-RT agent.

    
\section{Performance Evaluation}\label{sec:eval}
In this section, we evaluate the performance of our proposed agentic AI framework and compare it with three baseline schemes.
The considered cell-free O-RAN consists of $L=50$ O-RUs deployed in a $500 \textrm{ m}^2$ area. 
Each user is served by at most $L^\textrm{max}=8$ O-RUs.
The maximum transmit power of each O-RU $P^{\textrm{max}}$ is set to $30$ dBm.
Each O-RU and user device are equipped with $N_{\textrm{t}}=4$ and $N_{\textrm{r}}=2$ antennas, respectively.
We set the number of data streams $N_\textrm{s}=\min(N_\textrm{t}, N_\textrm{r})=2$.

For the proposed DRL algorithm, the discount factor $\gamma$ is set to $0.9$. The episode length $T$ is $10$. The clipping range $\epsilon$ is $0.2$. The policy and critic learning rates, $\delta_l$ and $\delta_v$, are both $10^{-4}$. 
We use GPT-5 via OpenAI's API as both the supervisor agent and the teacher model for the near-RT agents. We use a developer message to specify the expected behavior and output format.
For the lightweight student model, we deploy Qwen 2.5 \cite{LLM:Qwen2.5} with 7B and 14B parameters and compare the performance across the two model sizes. The adapter rank and the scaling parameter are $(d_\textrm{r}, \eta) = (32, 64)$ and $(64, 128)$ for the two model sizes, respectively.

\begin{table}[t]
  \centering
  \caption{Comparison of memory usage for the three near-RT agents under different quantization and LoRA deployment settings}
  \resizebox{\linewidth}{!}{
  \begin{tabular}{|c|c|c|c|c|}
    \hline
    \multirow{2}{*}{Model Size} & \multirow{2}{*}{$3\times$ FP16 LLMs} & $1\times$ FP16 LLM + & \multirow{2}{*}{$3\times$ FP4 LLMs} & $1\times$ FP4 LLM +\\
    &  & $3$ adapters &  & $3$ adapters\\
    \hline\hline
    7B & 45.7 GB & 15.3 GB & 11.4 GB & \textbf{3.8 GB}\\
    \hline
    14B & 88.2 GB & 29.5 GB & 22.1 GB & \textbf{7.4 GB}\\
    \hline\hline
    Reduction in & \multirow{2}{*}{-} & \multirow{2}{*}{67 \%} & \multirow{2}{*}{75 \%} & \multirow{2}{*}{\textbf{92 \%}}\\
    Memory Usage & & & & \\
    \hline
  \end{tabular}
  }
  \label{tab:QLoRA_size}
\end{table}

\begin{figure}[t]
\centering
{\subfloat[]{\includegraphics[width=0.49\linewidth]{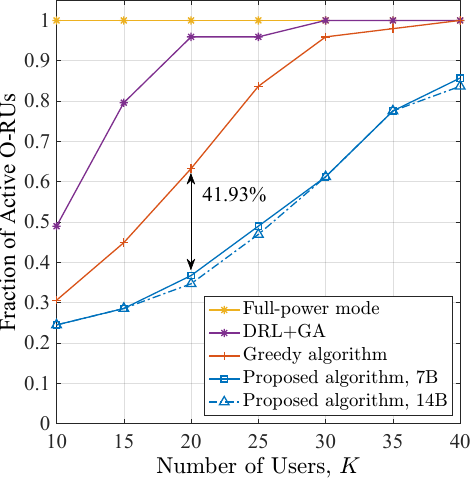}}}
{\subfloat[]{\includegraphics[width=0.49\linewidth]{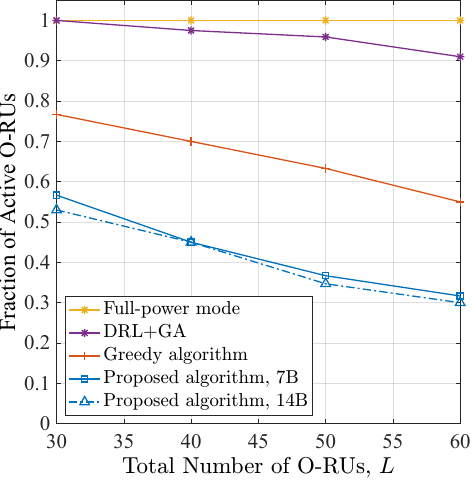}}}
\caption{The fraction of active O-RUs versus (a) the number of users (b) the total number of O-RUs}
\label{fig:active_ORUs}
\end{figure}

We consider three O-RU activation baselines. In the first baseline, referred to as DRL+gradient ascent (GA), we use the proposed DRL algorithm but simultaneously update $\mu_k$ and $\lambda_k$ as $\mu_k \leftarrow \mu_k + \delta_\mu \upsilon_k$ and $\lambda_k \leftarrow \lambda_k + \delta_\lambda \upsilon_k$, respectively, where $\delta_\mu$ and $\delta_\lambda$ are the step sizes. The second baseline is a greedy algorithm. For each user $k$, it activates the O-RU with the largest $\beta_{k,l}$. If the user’s minimum rate requirement is not satisfied, it incrementally activates the O-RU with the next largest $\beta_{k,l}$. The third baseline corresponds to a full-power mode, where all O-RUs remain active.

First, we evaluate the performance of the proposed framework in the energy-saving mode with $R_k^\textrm{min}=10 \textrm{ Mbps},\; k \in \mathcal{K}$. Fig. \ref{fig:active_ORUs}(a) shows the fraction of active O-RUs versus the number of users for the baselines and the proposed framework with different LLM model sizes. 
The proposed framework achieves similar performance with both 7B and 14B models and outperforms the greedy algorithm by up to $41.93\%$. In the DRL+GA case, the lack of coordination between the user weighting and O-RU management agents causes instability. The coefficient $\lambda_k$ quickly grows before the user weights $\alpha_k$ converge, which forces the DRL algorithm to activate many O-RUs to meet the minimum data rate requirements.

Fig. \ref{fig:active_ORUs}(b) shows the fraction of active O-RUs versus the total number of O-RUs with $K = 20$ users. As $L$ increases, the fraction of active O-RUs decreases for both the baselines and the proposed framework. The proposed framework consistently outperforms the baseline schemes.

\begin{table}[t]
  \centering
  \caption{Two examples of operator intents and the corresponding supervisor messages to the near-RT agents}
  \resizebox{\linewidth}{!}{
  \begin{tabular}{|c|c|c|}
    \hline
    Intent Type & Energy Saving (ES) & Utility Maximization (UM)\\
    \hline\hline
    \multirow{2}{*}{Operator} & Enter the energy-saving mode. & Maximize the sum of log-rates.\\
    & Guarantee 50 Mbps for user 3. & No minimum rate requirements.\\
    \hline
    Supervisor to & Objective: $\sum_{k \in \mathcal{K}} r_k$ & \multirow{2}{*}{Objective: $\sum_{k \in \mathcal{K}} \log (r_k)$}\\
    User Weighting& Constraint: $r_3 \geq 50 \textrm{ Mbps}$ & \\
    \hline
    Supervisor to & Objective: Energy Saving & \multirow{2}{*}{Objective: Full Power}\\
    O-RU Management& Constraint: $r_3 \geq 50 \textrm{ Mbps}$ & \\
    \hline
    Supervisor to & \multirow{2}{*}{Monitor: $r_3 \geq 50 \textrm{ Mbps}$} & \multirow{2}{*}{-}\\
    Monitoring & &\\
    \hline
  \end{tabular}
  }
  \label{tab:intent_example}
\end{table}

\begin{figure}[t]
\center{\includegraphics[width=0.65\linewidth]{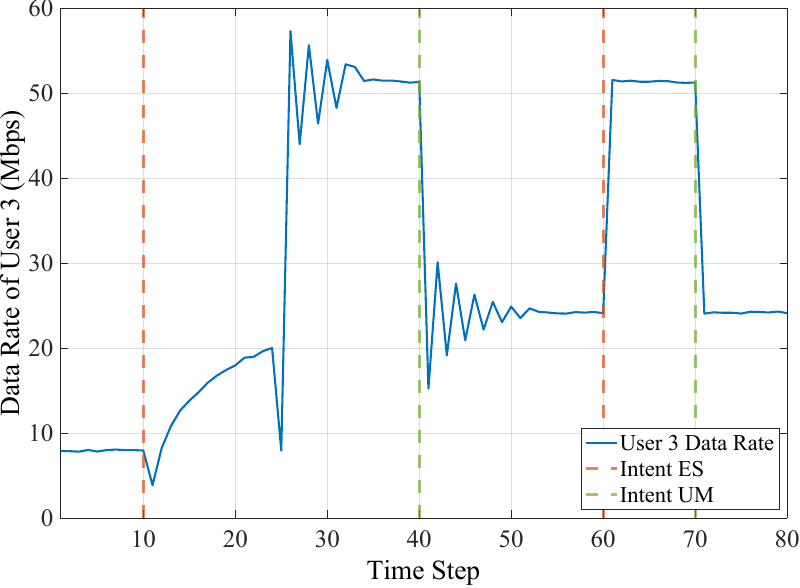}}
\caption{The data rate of user 3 for different operator intents.}
\label{fig:rate_convergence}
\end{figure}

Table \ref{tab:QLoRA_size} compares the total memory usage of the near-RT agents under different quantization and LoRA deployment settings. By jointly applying quantization and LoRA, the proposed framework reduces the overall memory usage by approximately 92\% across both model sizes when compared with deploying three separate full-precision LLMs.

Table \ref{tab:intent_example} provides examples of the energy-saving (ES) and utility-maximization (UM) intents and the corresponding objectives extracted by the supervisor agent. Fig. \ref{fig:rate_convergence} shows how the data rate of user 3, $r_3$, changes as different intents are applied. At $t=10$, the ES intent is applied. Consequently, the O-RU management agent deactivates several O-RUs and $r_3$ decreases. The monitoring agent then prompts the user weighting agent to increase $\alpha_3$ until the minimum rate requirement of user $3$ is satisfied. At $t=24$, the monitoring agent realizes that $r_3$ is converging to a point below $R_3^\textrm{min}$ and prompts the O-RU management agent to increase $\lambda_3$ and reactivate nearby O-RUs. The user weighting agent is then prompted again to adjust $\alpha_k$ until $r_3$ converges to $R_3^\textrm{min}$. At $t=40$, when the UM intent is applied, all O-RUs are activated. Since no minimum rate requirements are specified, the user weighting agent sets $\mu_k=0,\; k \in \mathcal{K}$ and updates $\alpha_k=\tilde{U}'_k(r_k)=\frac{1}{r_k}$ until the data rates stabilize. The next time the operator applies either ES or UM intents, the searching phase is skipped and the previously stored coefficients are directly retrieved from the memory.
\section{Conclusion}\label{sec:conclusion}
In this paper, we proposed an agentic AI framework for intent-driven optimization in cell-free O-RAN. A supervisor agent in the non-RT RIC translates operator intents into an objective function and minimum rate requirements. Based on this information, the user weighting agent in the near-RT RIC retrieves relevant prior experience from the memory to determine the user priority weights. The O-RU management agent uses a multi-agent DRL algorithm to determine the set of active O-RUs. The monitoring agent measures and monitors user data rates and coordinates with other agents to ensure that the minimum rate requirements are satisfied. To enhance scalability, we deployed a lightweight LLM in the near-RT RIC and trained a QLoRA adapter for each near-RT agent. Simulation results showed that the proposed framework reduces the number of active O-RUs by $41.93\%$ when compared with three O-RU sleeping baselines. It also reduces memory usage by $92\%$ when compared with deploying separate LLM agents. 
For future work, we plan to introduce additional agents for resource block allocation and channel estimation.

\bibliographystyle{IEEEtran}
\bibliography{Reference}
\end{document}